\documentclass[11pt]{article}
\usepackage{acl}
\usepackage{times}
\usepackage{latexsym}
\usepackage[T1]{fontenc}
\usepackage[utf8]{inputenc}
\usepackage{microtype}
\usepackage{inconsolata}
\usepackage{tabularx} 
\usepackage{graphicx}
\usepackage{tcolorbox}
\tcbuselibrary{skins} 
\usepackage{array}    
\usepackage{graphicx}   
\usepackage{booktabs}  
\usepackage{multirow}   
\usepackage{colortbl}   
\usepackage{xcolor}     
\usepackage{tabularx}   
\usepackage{diagbox}
\usepackage{amsfonts}
\usepackage{amssymb}
\usepackage{arydshln}
\usepackage{xspace,mfirstuc,tabulary}
\usepackage{amsmath}  
\usepackage{amssymb}  
\usepackage{dcolumn}
\usepackage{enumitem}
\newcolumntype{d}{D{.}{.}{2.2}}
\usepackage{tablefootnote}

\title{What to Format and How:
A Benchmark and Workflow Approach for Document Formatting}

\usepackage{xspace}
\newcommand{\docformbench}{\textsc{DocFormBench}\xspace}
\newcommand{\docformflow}{\textsc{DocFormFlow}\xspace}

\author{Shihao Rao$^{1, 2}$, Liang Li$^{1}$\thanks{$^{}$Corresponding author: Liang Li}, Jiapeng Liu$^{1, 2}$, Tong Lin$^{1, 2}$, \textbf{Bing Li}$^1$,\\
\textbf{Xiyan Gao}$^1$, \textbf{Peng Fu}$^1$,\textbf{Jing Huang}$^1$, \textbf{Can Ma}$^1$\\
$^1$Institute of Information Engineering, Chinese Academy of Sciences, Beijing, China \\
$^2$School of Cyber Security, University of Chinese Academy of Sciences, Beijing, China\\
\texttt{\{raoshihao, liliang\}@iie.ac.cn} \\
}

\begin{document}
\maketitle
\begin{abstract}
Recent advances in large language models (LLMs) have opened up new possibilities for automated document formatting. 
However, real-world formatting often requires identifying targets based on document content. This content-aware setting remains challenging and underexplored, primarily due to the lack of dedicated evaluation datasets.
To enable evaluation in realistic content-aware scenarios, we introduce DocFormBench, a benchmark that extends Text-to-Format evaluation to diverse formatting requirements, along with metrics for both accuracy and efficiency.
To mitigate redundant document reading in existing methods during formatting, we propose DocFormFlow, a workflow formatting method that decouples target localization from modification execution into \textit{what to format} and \textit{how}.
Extensive experiments across multiple LLMs and multimodal models show that DocFormFlow consistently improves formatting accuracy while reducing token consumption compared to representative baselines.
Further analysis reveals that precise target localization is the primary factor influencing formatting performance.
We hope DocFormBench and DocFormFlow will facilitate future research toward more intelligent and reliable document formatting.\footnote{We will make our code and dataset publicly available upon the acceptance of this paper.}

\end{abstract}

\section{Introduction}
\label{sec:introduction}

\begin{figure}[t] 
\centering 
\includegraphics[width=0.5\textwidth]{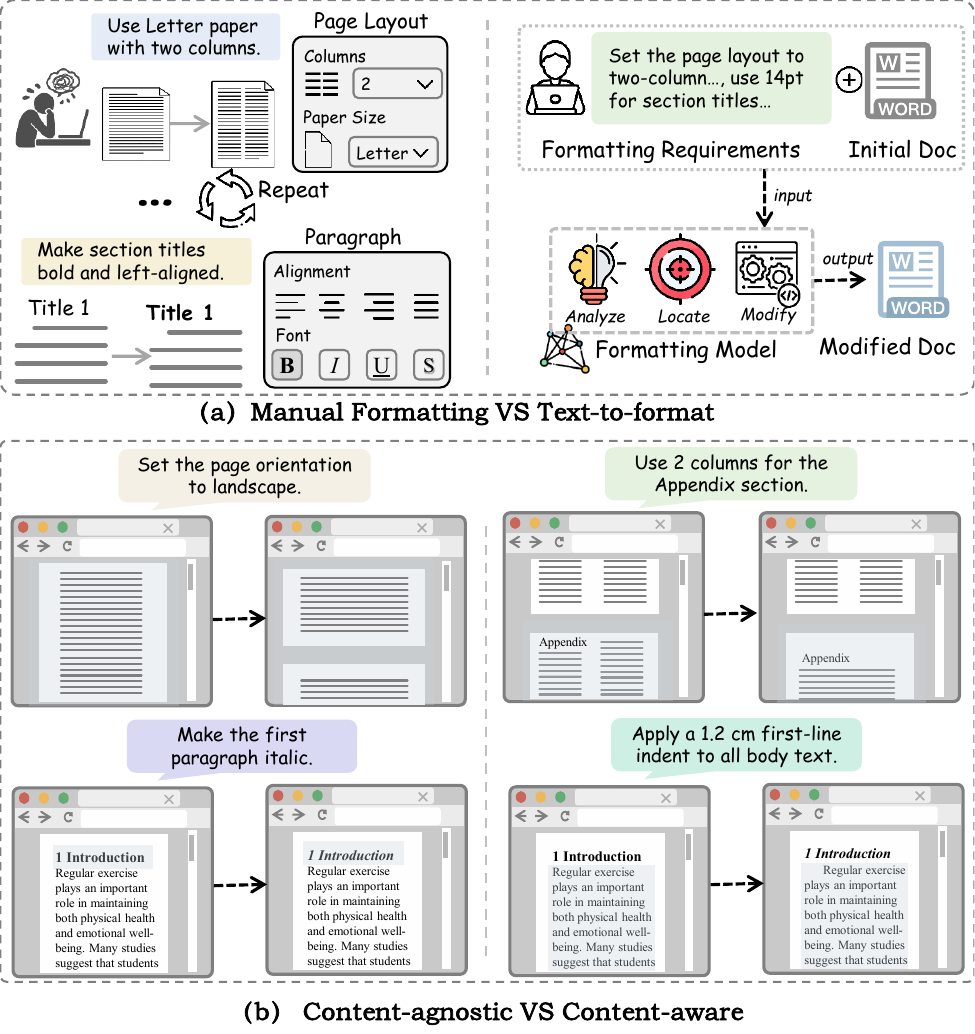} 
\caption{Overview of the Text-to-Format task, comparing manual and automated formatting as well as different types of formatting requirements.} 
\label{fig:overview} 
\end{figure}

Document formatting is a cornerstone of professional document creation, as it affects readability, visual consistency, and compliance with domain-specific standards~\citep{schwen2020ten,format-important}. Modern word processors such as Microsoft Word provide a wide range of formatting tools, yet applying formatting changes to long or structurally complex documents remains time-consuming and labor-intensive~\citep{leblanc2019scientific}.
To reduce this burden, as illustrated in Figure~\ref{fig:overview} (a), the \textbf{Text-to-Format} task aims to translate natural-language formatting requests into executable document modifications~\citep{rao-etal-2024-free, wu2024agentdocedit},
involving deciding \textit{what to modify} and then executing \textit{how to modify it} across multiple elements in the same document.

As illustrated in Figure~\ref{fig:overview} (b), real-world formatting requests generally fall into two broad categories. 
The first is \textit{content-agnostic}, which can be fulfilled without interpreting document semantics, such as globally adjusting line spacing or page layout. The second is \textit{content-aware}, which requires semantic understanding to identify the elements referred to in natural language, such as bolding all section titles. 
Accordingly, early approaches~\citep{docxtemplates, xkonglong_gw, rao-etal-2024-free} address content-agnostic requests by operating on templates, regular expressions, or manually specified regions, without interpreting document content. 
More recent methods, such as GUI Agents~\citep{ufo2, AgentS2} and DocxSkills~\citep{anthropic_docx_skill}, handle content-aware formatting by understanding documents to locate targets and apply formatting.
They typically employ ReAct-style frameworks~\citep{yao2022react, kim2025reflact} that interleave reasoning, action, and observation, decomposing tasks into sub-tasks for step-by-step execution.

Despite these advances, existing work on Text-to-Format still faces two major limitations.
For evaluation, existing benchmarks lack dedicated support for content-aware formatting scenarios. DocFormEval~\citep{rao-etal-2024-free} focuses primarily on content-agnostic settings, while general-purpose GUI benchmarks like OSWorld~\citep{osworld} include only a handful of formatting tasks.

Furthermore, we observe that existing ReAct-style methods tightly couple target localization and formatting execution at each step, requiring repeated full document reviewing before every modification.
This step-by-step re-localization leads to redundant reasoning, resulting in high token consumption and potential inefficiency.

To address these limitations, we first introduce \docformbench, a benchmark that extends evaluation to realistic content-aware formatting scenarios, filling the gap left by prior datasets.
It comprises 500 carefully verified instances across 12 widely used document types in both Chinese and English, spanning diverse formatting requirements.
Additionally, we propose four evaluation metrics, namely Format Accuracy, Hallucination Formatting Rate, Average LLM Calls, and Average Tokens consumption per LLM call, to comprehensively assess the formatting method across three dimensions: formatting accuracy, formatting scope precision, and execution efficiency.

To mitigate repeated document reviewing before each modification, we introduce \docformflow, a workflow formatting method that separates target identification from modification execution.
 It first identifies all target elements, then performs the required formatting in a subsequent stage, keeping each step focused on a single responsibility. 
Concretely, \docformflow consists of four stages: requirement expansion, intent classification, target element localization, and verified format modification. The first three stages determine \textit{what to modify}, while the final stage handles \textit{how to modify it}, ensuring a clean separation between formatting intent analysis and modification execution.

In summary, our contributions are threefold:
\begin{itemize}[nosep]
    \item We introduce \docformbench, a benchmark for content-aware Text-to-Format tasks that enables systematic evaluation and advances research toward more realistic document formatting scenarios.

    \item We propose \docformflow, a workflow formatting method which structures document formatting as an orderly two-stage process. This separation keeps each stage focused and avoids repeatedly reviewing the entire document for localization.

    \item We conduct extensive experiments on \docformbench, showing that \docformflow achieves the best formatting accuracy with lower token consumption, while highlighting the importance of accurate target localization for efficient document formatting.
    
\end{itemize}

\section{Related Work}

\paragraph{Content-Agnostic Formatting Methods}
Document formatting can often be performed using templates, positional rules, or deterministic patterns, without requiring semantic understanding of document content.
Representative approaches include template-driven document generation systems~\citep{docxtemplates, xkonglong_gw} and code-generation-based formatting methods~\citep{rao-etal-2024-free}, both of which operate on predefined target specifications.
While effective in structured scenarios where formatting targets are explicitly defined, these methods cannot accommodate instructions that depend on document content to identify target elements, limiting their applicability to realistic document formatting tasks.

\paragraph{Content-Aware Formatting Methods}
To support content-dependent formatting, recent methods incorporate document understanding to identify target elements from natural-language instructions.
Most existing approaches are adapted from general-purpose document agents rather than designed specifically for formatting, and can be broadly categorized by how formatting actions are executed: GUI-based interaction and API-based invocation.

GUI-based methods~\citep{ufo2, AgentS2, Ui-tars-2, seagent} perform formatting by interacting with document editors, typically using large language models (LLMs) to interpret interface elements.
These methods can leverage screen information to assist element localization and formatting, and execute actions through simulated human interactions.

While the interaction-based execution enables broad applicability across software environments, it is often computationally expensive for long sequences of repetitive operations and may be vulnerable to cascading failures caused by localization errors~\citep{osworld, nguyen2025gui,windows-use}.

API-based methods~\citep{autogen, autotool, openclawrl2026} instead invoke document manipulation interfaces directly to apply formatting changes.
Systems such as Docx-Skills~\citep{anthropic_docx_skill} and OfficeCLI~\citep{officecli} expose formatting operations as reusable tools within reasoning-action loops, improving execution efficiency and precision.

However, these ReAct-style methods may suffer from structural mismatch with the iterative locate-and-modify pattern required by Text-to-Format tasks.
They may repeatedly review document content and execute adjustment action, potentially leading to redundant reasoning overhead and low accuracy on Text-to-Format tasks.

\section{\docformbench Dataset}

\subsection{Task Formulation}
\label{sec:formulation}
The document formatting task can be defined as follows: given a source Word document $D_{src}$ and a natural-language formatting request $R$ describing the formatting requirement, the model is expected to generate a target document $D_{tgt}$ that satisfies the specified requirement. Formally,
\[\mathcal{M}: (D_{src}, R) \mapsto D_{tgt}\]

where $\mathcal{M}$ denotes a formatting model that typically involves document reading, instruction understanding, target element localization, and formatting execution.

\subsection{\docformbench Construction Pipeline}
\label{sec:pipeline}

\begin{figure}[t]
    \centering
    \includegraphics[width=0.5\textwidth]{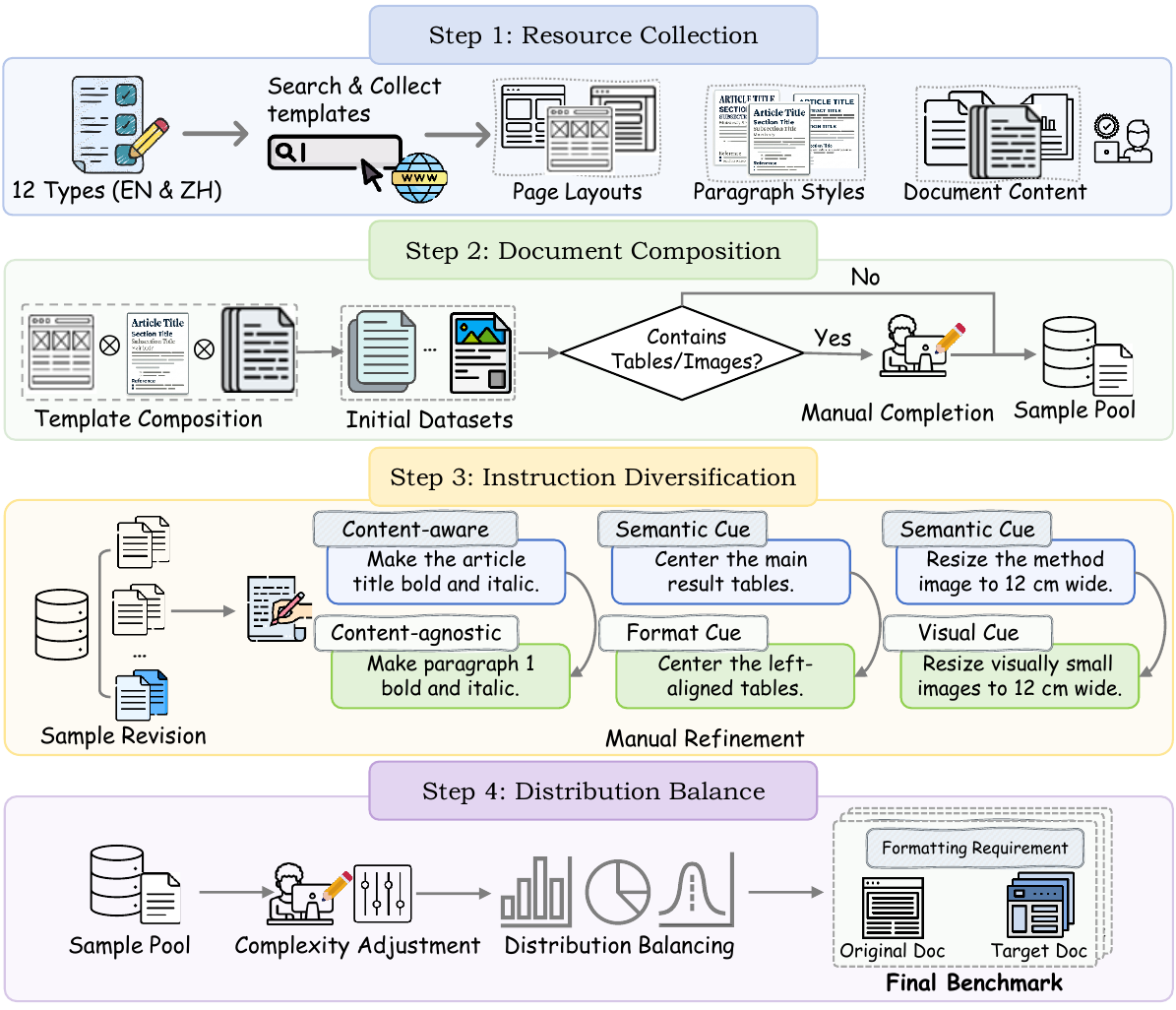}
    \caption{
    Process of the benchmark construction, including resource collection, document composition, instruction diversification, and distribution balancing.
    }
    \label{fig:pipeline}
\end{figure}

To ensure alignment with the design goals of DocFormBench, we construct the dataset following three core principles: realism, diversity, and controllability. 
Realism ensures that samples reflect real-world document formatting scenarios, diversity ensures broad coverage across document types, languages, and instruction expressions, and controllability enables systematic composition for reliable evaluation.
To achieve these goals, as illustrated in Figure~\ref{fig:pipeline}, we propose a four-stage pipeline that combines automated generation with manual refinement, consisting of:

\paragraph{Resource Collection}
It is observed that in real-world scenarios, document formatting styles vary significantly across different languages and document types, while formatting requests naturally include both \textit{content-agnostic} and \textit{content-aware} cases.
Motivated by these observations, we first construct foundational resources along two dimensions: document type (e.g., academic papers, government documents, legal files) and language (Chinese and English).
This structured design enables unified modeling of both content-agnostic and content-aware formatting scenarios within a single framework.

For each combination $(t, l)$, we define three types of resources:
\begin{itemize}[nosep]
    \item \textbf{Page Layout Templates} $\mathcal{P}_{t,l}$: specify page-level formatting rules, corresponding to content-agnostic formatting operations.
    \item \textbf{Paragraph Templates} $\mathcal{S}_{t,l}$: define semantic or structural paragraph styles, enabling content-aware formatting based on document roles.
    \item \textbf{Document Content} $\mathcal{C}_{t,l}$: textual content annotated with semantic roles, providing grounding signals for content-aware formatting.
\end{itemize}

This decomposition explicitly separates layout, style, and content, enabling flexible recombination while preserving control over formatting factors. More details are provided in Appendix~\ref{app:resource_details}.

\paragraph{Document Composition}

Given each pair $(t, l)$, we generate documents through systematic composition of layout templates, paragraph styles, and content blocks.
Specifically, each document is constructed as:
\[
    \mathcal{D}_{\text{init}}^{t,l} = \bigl\{ \Phi(p, s, c) \mid p \in \mathcal{P}_{t,l}, s \in \mathcal{S}_{t,l}, c \in \mathcal{C}_{t,l} \bigr\},
\]
where $\Phi$ denotes the rule-based composition that applies paragraph styles to textual content and enforces page-level layout.

This compositional design enables scalable generation of diverse formatting instances.
In addition, to better cover complex real-world scenarios, particularly those involving tables and images, we further introduce a manual completion stage. This step addresses the limited coverage of non-textual formatting scenarios in automatic composition, ensuring that complex tasks involving tables and images are adequately represented.

\paragraph{Instruction Diversification}
Although the initial instructions generated during document composition cover diverse formatting operations, their referring expressions are often overly templated and do not fully reflect how users naturally specify target elements in practice.
To improve realism, we rewrite only the referring expressions used to identify the document elements to be modified, while keeping the intended formatting operations unchanged, as shown in Figure \ref{fig:pipeline}.
These rewrites are guided by the specific document context, ensuring that the resulting instructions more closely resemble natural formatting requests.
This process increases expression diversity and improves the realism of instruction formulations without altering their underlying formatting intent.

\paragraph{Distribution Balance}
The initial construction process tends to produce an uneven distribution of difficulty, with some samples requiring overly complex formatting.
To obtain a more balanced benchmark, we manually adjust the sample distribution across document length, topic category, and formatting operation type.
Specifically, we simplify some overly complex samples by removing unnecessary formatting requirements and discarding a small number of exceptionally difficult cases.
These adjustments lead to a more realistic and balanced complexity distribution, better reflecting the range of formatting tasks encountered in real-world scenarios.
The resulting dataset $\mathbb{D}_{\text{final}}$ provides a balanced evaluation benchmark across diverse formatting scenarios and difficulty levels.

From the above pipeline, we construct DocFormBench.
Each sample in $\mathbb{D}_{\text{final}}$ is represented as a triplet $\langle D_{\text{src}}, R, D_{\text{tgt}} \rangle$, where $D_{\text{src}}$ is the source document, $R$ is the natural language formatting requirement, and $D_{\text{tgt}}$ is the target document satisfying the formatting requirement.
Due to page limitations, we provide a statistical analysis of \textbf{DocFormBench} in Appendix \ref{app:statistic}. Please refer to it for more details.

\section{DocFormFlow}
\label{sec:method}
\begin{figure*}[t]
    \centering
    \includegraphics[width=\textwidth]{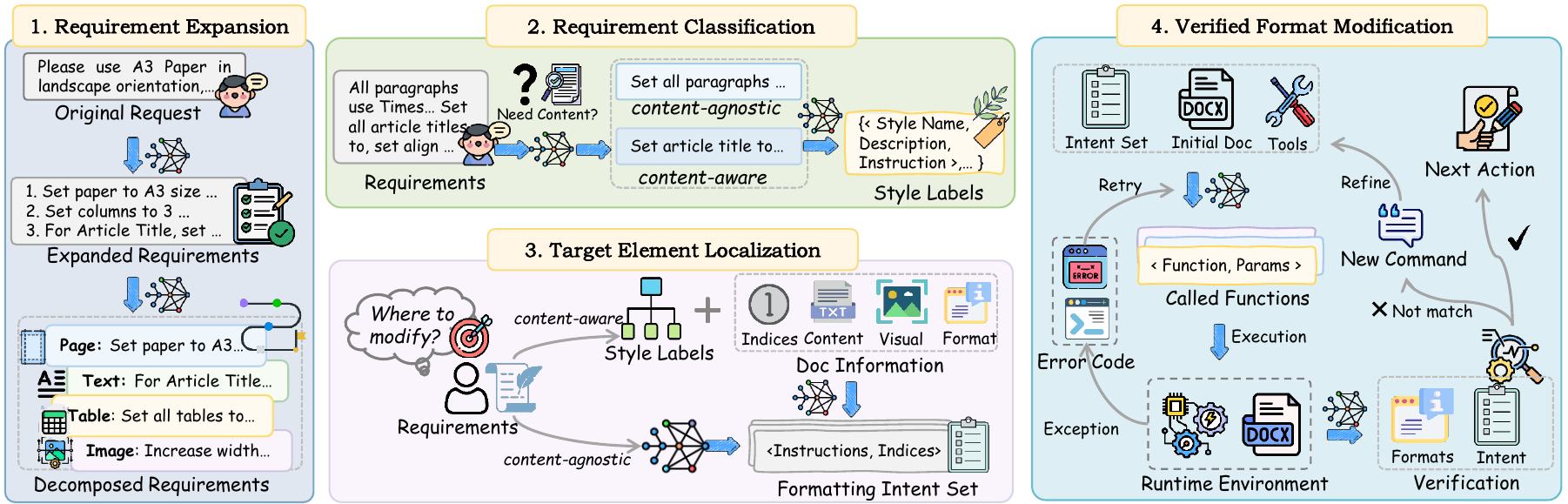}
    \caption{The architecture of our proposed \docformflow.}
    \label{fig:method}
\end{figure*}

Prior work~\citep{metatool, apicalls, autotool} shows that large language models (LLMs) can translate natural language instructions into executable operations via instruction understanding and function calling. Inspired by this, we propose \docformflow, a workflow approach that separates \textit{what to modify} and \textit{how}, decomposing the formatting process into four stages. 
As illustrated in Figure~\ref{fig:method}, the process proceeds in four stages: (1) globally identifying the target objects, (2) for each object, classifying how the targets are specified, (3) resolving their precise document positions, and (4) executing verified formatting operations. We detail each stage as follows:

\subsection{Requirement Expansion}
The first stage determines \textit{which objects} to modify.
Formatting requests are often incomplete: some target objects may be omitted, and requirements may be expressed only partially or implicitly.
To better capture the formatting intent, we first expand the original request \(R\) into a more explicit requirement \(R_{\text{ex}}\) using an LLM \(\mathcal{M}\):
\[
\mathcal{M}: R \mapsto R_{\text{ex}}.
\]
Then, prior work shows that decomposing complex tool-use tasks reduces complexity and improves reliability, we decompose \(R_{\text{ex}}\) into object-specific sub-requirements:
\[
\mathcal{M}: R_{\text{ex}} \mapsto \{R_{\text{page}},\, R_{\text{text}},\, R_{\text{table}},\, R_{\text{image}}\}.
\]
Each sub-requirement is then handled independently through the remaining stages of classification, localization, and verified modification.

\subsection{Requirement Classification}
Once target objects are identified, we determine how their locations are specified. Formatting requirements differ in whether they depend on semantic understanding of document content (see Section~\ref{sec:introduction}). Using an LLM \(\mathcal{M}\), we classify each sub-requirement:
\[
\mathcal{M}: R_{\text{obj}} \mapsto \{R_{\text{agnostic}},\, R_{\text{aware}}\},
\]
where \(R_{\text{agnostic}}\) is content-agnostic and \(R_{\text{aware}}\) is content-aware.

A single content-aware requirement may refer to multiple elements. We therefore decompose \(R_{\text{aware}}\) into a set of element-level tuples:
\[
R'_{\text{aware}} = \mathcal{M}(R_{\text{aware}}) = \{\langle s_i, u_i, d_i \rangle\}_{i=1}^{k},
\]
where \(s_i\) is a style label (a unique category assigned by the LLM to the target element), \(u_i\) the formatting instruction, and \(d_i\) a textual description for downstream localization. Each tuple represents one target element, enabling fine-grained grounding in the next stage.

\subsection{Target Element Localization}
\label{sec:target-localization}

Executing formatting instructions requires precise element indices in the document. We therefore convert the requirements \(R_{\text{agnostic}}\) and \(R'_{\text{aware}}\) into a unified set of execution mappings \(\mathcal{S} = \mathcal{S}_{\text{agnostic}} \cup \mathcal{S}_{\text{aware}}\), where each mapping \(\langle u_i, \mathcal{P}_i \rangle\) pairs a formatting instruction with its target indices.

For content-agnostic requirements, the LLM directly produces \(\langle u_i, \mathcal{P}_i \rangle\) from \(R_{\text{agnostic}}\):
\[
\mathcal{M}: R_{\text{agnostic}} \mapsto \{ \langle u_i, \mathcal{P}_i \rangle \}.
\]

For content-aware requirements, we represent each candidate document element as \(c_j = \langle \mathrm{idx}_j, \mathrm{f}_j, \mathrm{t}_j, \mathrm{img}_j \rangle\), where \(\mathrm{idx}_j\) is its index, \(\mathrm{f}_j\) its formatting, \(\mathrm{t}_j\) its surrounding text, and \(\mathrm{img}_j\) a rendered page image. A locator \(\mathcal{L}\) then maps the element-level tuples \(\langle s_i, u_i, d_i \rangle\) to indices by classifying each \(c_j\) into one of the style labels \(s_i\) or “other”:
\[
\mathcal{L}\big( \{\langle s_i, u_i, d_i \rangle\}, \{c_j\} \big) \mapsto \{ \langle u_i, \mathcal{P}_i \rangle \}.
\]

Combining mappings from both categories yields the complete execution set \(\mathcal{S}\), specifying precisely \textit{what to format} for the downstream modification stage.

\subsection{Verified Format Modification}
\label{sec:verified-modification}
Given the execution set \(\mathcal{S} = \{ \langle u_i, \mathcal{P}_i \rangle \}\) from the previous stage, this stage performs the formatting operations, addressing \textit{how to modify} the document. Since LLM-generated function calls may contain bugs or misalign with human intent~\cite{critictool, funcallerror}, we introduce a verification mechanism for robustness and accuracy.

Each mapping \(\langle u_i, \mathcal{P}_i \rangle\) is processed sequentially. The LLM generates a function call to the appropriate formatting tool and executes it. If an exception occurs, the error is fed back for revision and retry. On success, the updated formatting attributes are read and verified against \(u_i\). If verification passes, the system proceeds to the next mapping; otherwise, the LLM produces a corrected instruction \(\tilde{u}_i\) and retries. This loop continues until acceptance or a maximum iteration limit is reached.

\section{Experimental Setup}

\subsection{Baselines}

We evaluate \docformbench against four representative approaches that tackle content-aware Text-to-Format requests. According to their interaction paradigms, these baselines fall into two categories: \textsc{GUI-Base} and \textsc{API-Base}. 

To ensure fair comparison and easy reproducibility, we select only publicly available baselines that require no external grounding models. 
GUI-Base methods simulate human interactions with the graphical interface. 
We select easy-to-easy representatives: 
\begin{itemize}[nosep]
\item {\textsc{Windows-Use}}~\citep{windows-use}: A lightweight open-source GUI agent that leverages Windows UI Automation to parse interface elements and supports screen input to guide formatting.
\item \textsc{UFO2}~\citep{ufo2}: A multi-agent desktop AgentOS designed for Windows. It can perform document formatting through simulated interactions and, optionally, native API calls.
\end{itemize}
API-Base methods rely on programming interfaces for formatting control.
We select methods specifically designed for Word document operations:
\begin{itemize}[nosep]
\item \textsc{Docx‑Skill}~\citep{anthropic_docx_skill}: An open-source document editing and modifying skill for Word documents. In our experimental setup, it is invoked through the lightweight agent framework \texttt{nanobot}~\citep{nanobot}.
\item \textsc{MCP‑MSOffice}~\citep{msoffice}: An open-source MCP tool that exposes Microsoft Word operations on Windows, enabling formatting tasks through structured API calls.
\end{itemize}

\subsection{Evaluation Metrics}

To comprehensively evaluate formatting method performance, we propose four metrics as follows:
(1) Format Accuracy (FA): the proportion of formatting attributes that exactly match the target requirements.

(2) Hallucination Formatting Rate (HFR): the proportion of formatting modifications applied outside the intended scope. A lower value indicates better precision.

(3) We use two metrics to evaluate the execution efficiency of the formatting method: the average number of LLM calls (AVG Calls) and token consumption per LLM call (AVG Tokens).
Detailed calculations are provided in Appendix~\ref{app:metrics}.

\subsection{Implementation Details}

We evaluate all methods and models on \docformbench using both multimodal and large language models, as listed in Table~\ref{tab:model_list}. 
All models are accessed via publicly available APIs, with reasoning (thinking) mode disabled by default and enabled only when analyzing its effect. 
For \textsc{DocFormatFlow}, we set the temperature to 1.0, other hyperparameters to defaults, and limit internal format self-repair and verification to 3 iterations. 
Baseline methods use their default hyperparameters, and GUI-Agent baselines are evaluated exclusively on multimodal models due to their reliance on visual grounding. 
To ensure computational consistency, we cap inference steps at $\textit{max\_step}=50$ and implement a mechanism to prevent infinite loops.
Detailed settings and strategies are in Appendix~\ref{app:param_settings}.

\begin{table}[!t]
  \centering
  \resizebox{\columnwidth}{!}{%
    \begin{tabular}{llrrrr}
      \toprule
      \textbf{Model} & \textbf{Method} & \textbf{AVG Calls} & \textbf{FA(\%)} & \textbf{HFR$\downarrow$} & \textbf{AVG Tokens$\downarrow$} \\
      \midrule
      \rowcolor{gray!20} \multicolumn{6}{l}{\textit{Multimodal Models}} \\
      \multirow{5}{*}{GPT-5}
        & \textsc{Windows-Use} & 45.02 & 13.79 & 9.49 & 14966 \\
        & \textsc{UFO-2} & 45.40 & 16.91 & 2.07 & 26165 \\
        & \textsc{MCP-Msoffice} & 4.37 & 12.32 & 0.88 & \textbf{2651} \\
        & \textsc{Docx-Skill} & 4.00 & 4.49 & \textbf{0.43} & 5369 \\
        & \docformflow & 33.62 & \textbf{72.53} & 3.44 & 2998 \\
      \midrule
      \multirow{5}{*}{Gemini-3-Flash-Preview}
        & \textsc{Windows-Use} & 3.26 & 1.15 & 7.96 & 12466 \\
        & \textsc{UFO-2} & 42.10 & 8.42 & 9.05 & 23799 \\
        & \textsc{MCP-Msoffice} & 19.61 & 22.38 & 1.92 & 9413 \\
        & \textsc{Docx-Skill} & 4.67 & 23.55 & \textbf{0.95} & 8351 \\
        & \docformflow & 31.26 & \textbf{80.36} & 1.58 & \textbf{4187} \\
      \midrule
      \multirow{5}{*}{GLM-4.6v}
        & \textsc{Windows-Use} & 14.69 & 0.03 & 7.24 & 15056 \\
        & \textsc{UFO-2} & 2.69 & 0.21 & 8.05 & 37362 \\
        & \textsc{MCP-Msoffice} & 23.14 & 16.38 & 3.16 & 5536 \\
        & \textsc{Docx-Skill} & 10.31 & 25.57 & 4.16 & 11184 \\
        & \docformflow & 31.22 & \textbf{65.78} & \textbf{2.14} & \textbf{3765} \\
      \midrule
      \multirow{5}{*}{Qwen3-VL-235B-A22B-I}
        & \textsc{Windows-Use} & 6.10 & 0.13 & 9.49 & 13278 \\
        & \textsc{UFO-2} & 46.27 & 8.20 & 9.24 & 31816 \\
        & \textsc{MCP-Msoffice} & 13.49 & 12.11 & \textbf{1.93} & 6263 \\
        & \textsc{Docx-Skill} & 13.39 & 15.53 & 8.70 & 10819 \\
        & \docformflow & 29.28 & \textbf{70.94} & 2.03 & \textbf{3246} \\
      \midrule
      \rowcolor{gray!20} \multicolumn{6}{l}{\textit{Language Models}} \\
      \multirow{3}{*}{GLM-4.6}
        & \textsc{MCP-Msoffice} & 28.85 & 18.03 & 2.00 & 6284 \\
        & \textsc{Docx-Skill} & 39.17 & 18.19 & 3.07 & 17199 \\
        & \docformflow & 32.39 & \textbf{75.72} & \textbf{1.24} & \textbf{2659} \\
      \midrule
      \multirow{3}{*}{Qwen3-235B-A22B-I}
        & \textsc{MCP-Msoffice} & 24.74 & 14.44 & \textbf{1.99} & 5897 \\
        & \textsc{Docx-Skill} & 24.22 & 12.36 & 2.71 & 13372\\
        & \docformflow & 33.41 & \textbf{71.18} & 2.69 & \textbf{2829} \\
      \midrule
      \multirow{3}{*}{Deepseek-V3.2}
        & \textsc{MCP-Msoffice} & 30.02 & 15.41 & \textbf{1.61} & 9534 \\
        & \textsc{Docx-Skill} & 42.14 & 20.71 & 1.64 & 17074 \\
        & \docformflow & 34.77 & \textbf{72.87} & 1.99 & \textbf{2802} \\
      \bottomrule
    \end{tabular}
  }
  \caption{Performance comparison across models and methods. Metrics include AVG Calls, Accuracy (FA), Hallucination Formatting Rate (HFR), and Average Token Counts (AVG Tokens).}
  \label{tab:main_result}
\end{table}
\section{Experimental Results}

\subsection{Main Results}

Table~\ref{tab:main_result} presents the overall performance comparison across representative GUI-based and API-based baselines on \docformbench.

From the results, we draw four main observations as follows:

Firstly, \docformflow substantially improves formatting accuracy over existing baselines.
Across both multimodal and language-only models, \docformflow consistently achieves the highest FA, improving over the strongest baseline by large margins (e.g., +49.0\% on GPT-5 and +53.3\% on GLM-4.6).
In contrast, existing methods consistently fall below 25\% FA, revealing a difficulty in reliably locating and modifying intended targets.
GUI-based methods face the challenge of interface grounding and action execution, while API-based baselines suffer primarily at locating the correct target.
By explicitly separating localization and formatting, \docformflow reduces this ambiguity and enables more accurate formatting.

Secondly, the performance gain of \docformflow is robust across diverse model families.
Baseline performance varies substantially across different model backbones, especially for GUI-based methods, which often fail on weaker multimodal models.
In contrast, \docformflow maintains consistently strong accuracy (65.8\%--80.4\%) across all tested LLMs and MLLMs.
This stability suggests that the proposed workflow reduces dependence on model-specific agent capabilities, such as long-horizon planning or precise UI understanding.
This stability shows that \docformflow avoids heavy reliance on model-specific capabilities such as precise UI understanding, making it a more reliable solution for Text-to-Format tasks.

Thirdly, \docformflow achieves better effectiveness.
Although \docformflow requires multiple LLM interactions, its AVG Tokens remains much lower than GUI-based baselines and comparable to lightweight API-based methods.
For example, compared with \textsc{Windows-Use} and \textsc{UFO-2}, \docformflow reduces token usage by over 70\% while delivering substantially higher accuracy.
These results show that decoupling target localization from modification execution avoids repeatedly reading the entire document, thereby improving accuracy and reducing token consumption.

Finally, we notice that \docformflow does not always achieve the lowest HFR, i.e., it may occasionally over-modify elements beyond the intended scope. 
We consider this behavior to reflect the inherent limitations of current models in precisely localizing target elements.
In attempting to satisfy formatting requirements, the model tends to extend modifications to adjacent elements, leading to elevated HFR. 
Given its higher accuracy, this level of hallucination remains acceptable.

\begin{table}[!t]
  \centering
  \resizebox{\columnwidth}{!}{%
    \begin{tabular}{llllr}
      \toprule
      \textbf{Model} & \textbf{AVG Calls} & \textbf{FA (\%)} & \textbf{HFR $\downarrow$} & \textbf{AVG Tokens} \\
      \midrule
      \rowcolor{gray!20} \multicolumn{5}{c}{\text{Textual Input Only}} \\
      \rowcolor{gray!10} \multicolumn{5}{l}{\textit{Multimodal Models}} \\
      GPT-5                     & 36.52 & 72.32 & 3.03 & 2,998 \\
      Gemini-3-Flash-Preview    & 30.57 & 75.72 & 1.68 & 4,187 \\
      GLM-4.6v                  & 29.38 & 61.49 & 1.77 & 3,765 \\
      Qwen3-VL-235B-A22B-I      & 29.03 & 67.86 & 2.00 & 3,246 \\
      \rowcolor{gray!10} \multicolumn{5}{l}{\textit{Language Models}} \\
    Qwen3-235B-A22B-I         & 33.41 & 71.18 & 2.69 & 2,829 \\
      GLM-4.6                   & 32.39 & 75.72 & 1.24 & 2,659 \\
      \midrule
      \rowcolor{gray!20} \multicolumn{5}{c}{\text{Multimodal Input (Text + Visual)}} \\
      \rowcolor{gray!10} \multicolumn{5}{l}{\textit{Multimodal Models}} \\
      GPT-5                     & 33.62 & 72.53\textcolor{red}{$_{\uparrow 0.21}$} & 3.44\textcolor{red}{$_{\uparrow 0.41}$} & 2,792 \\
      Gemini-3-Flash-Preview    & 31.26 & 80.36\textcolor{red}{$_{\uparrow 4.64}$} & 1.58\textcolor{blue}{$_{\downarrow 0.10}$} & 4,056 \\
      GLM-4.6v                  & 31.22 & 65.78\textcolor{red}{$_{\uparrow 4.29}$} & 2.14\textcolor{red}{$_{\uparrow 0.37}$} & 2,870 \\
      Qwen3-VL-235B-A22B-I      & 29.28 & 70.94\textcolor{red}{$_{\uparrow 3.08}$} & 2.03\textcolor{red}{$_{\uparrow 0.03}$} & 2,779 \\
      \midrule
      \rowcolor{gray!20} \multicolumn{5}{c}{\texttt{Language Models with Visual Assistance$^\dagger$}} \\
      Qwen3-235B-A22B-I$^\dagger$ & 34.19 & 75.20\textcolor{red}{$_{\uparrow 4.02}$} & 2.24\textcolor{blue}{$_{\downarrow 0.45}$} & 3,301 \\
      GLM-4.6$^\dagger$           & 31.67 & 77.70\textcolor{red}{$_{\uparrow 1.98}$} & 1.42\textcolor{red}{$_{\uparrow 0.18}$} & 3,790 \\
      \midrule
      \rowcolor{gray!20} \multicolumn{5}{c}{\texttt{With Reasoning}} \\
      GPT-5                     & 34.53 & 74.62\textcolor{red}{$_{\uparrow 2.30}$} & 3.41\textcolor{red}{$_{\uparrow 0.38}$} & 3,553 \\
      Gemini-3-Flash-Preview    & 29.35 & 77.07\textcolor{red}{$_{\uparrow 1.35}$} & 1.59\textcolor{blue}{$_{\downarrow 0.09}$} & 4,314 \\
      Qwen3-235B-A22B-I         & 25.07 & 60.04\textcolor{blue}{$_{\downarrow 11.14}$} & 1.39\textcolor{blue}{$_{\downarrow 1.30}$} & 5,437 \\
      GLM-4.6$^\dagger$           & 31.67 & 77.70\textcolor{red}{$_{\uparrow 1.98}$} & 1.43\textcolor{red}{$_{\uparrow 0.19}$} & 5,093 \\
      \bottomrule
    \end{tabular}
  }
    \caption{Performance of \docformflow\ under different model configurations. FA represents Format Accuracy, HFR represents Hallucination Modification (lower is better). For rows with $\uparrow/\downarrow$, changes are measured relative to the corresponding \text{Textual Input Only} setting.}
      \label{tab:dorformflow_configs}
\end{table}

\subsection{Ablation Study}

We conduct ablation experiments to explore how different model configurations affect \docformflow. The results are shown in Table~\ref{tab:dorformflow_configs}, and we discuss three main findings below.

Firstly, \textbf{language-only models outperform multimodal counterparts.}
Under text-only input, language-only models consistently achieve higher FA and lower HFR than their multimodal versions within the same model family.
GLM-4.6 substantially exceeds GLM-4.6v in both FA and HFR, and a similar pattern appears for the Qwen3 models.
These results suggest that formatting performance depends strongly on semantic understanding for target localization and tool use.

Secondly, \textbf{visual input helps only when grounding is accurate.}
When visual signals are added, most models achieve higher formatting accuracy, suggesting that visual information can help locate more potential targets.
However, these gains often come together with higher HFR.
For example, GLM-4.6v and Qwen3-VL improve FA by over 3\%, but also produce more hallucinated modifications.
GPT-5 shows only a marginal FA gain (+0.21\%) while its HFR increases noticeably.
Only Gemini-3-Flash-Preview improves both FA and HFR, which may explain why it achieves the strongest overall performance.
These results suggest that visual input helps only when the model can accurately ground what it sees; without this, extra visual signals tend to trigger unnecessary modifications.

Lastly, \textbf{explicit reasoning adds cost without reliable improvement.}
Adding explicit reasoning yields modest FA gains for GPT-5 and Gemini, but also leads to much higher token consumption and sometimes higher HFR.
For Qwen3-235B-A22B-I, reasoning sharply reduces FA, even though HFR becomes lower, suggesting that the model becomes overly conservative rather than more accurate.
These results suggest that longer reasoning does not consistently improve formatting decisions.
Directly following instructions is often more reliable for balancing correct modifications and avoiding unnecessary changes.
These results suggest that explicit reasoning is often unnecessary once the formatting task is decomposed into simpler steps. 
Instead, additional reasoning may lead to overthinking and reduce formatting reliability.

\subsection{Further Analysis}

\begin{figure}[t]
    \centering
    \includegraphics[width=0.5\textwidth]{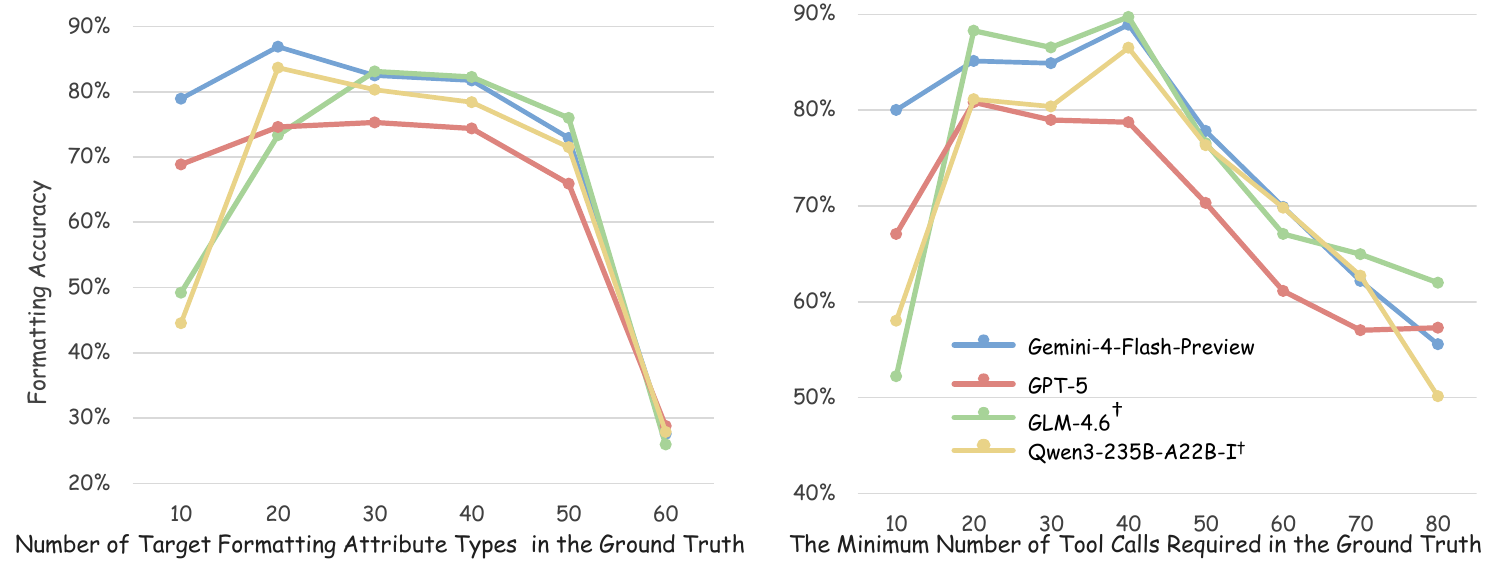}
    \caption{Impact of performance by formatting attribution category and number of tool calls required.}
    \label{fig:input_Influence}
\end{figure}

\begin{figure}[t]
    \centering
    \includegraphics[width=0.5\textwidth]{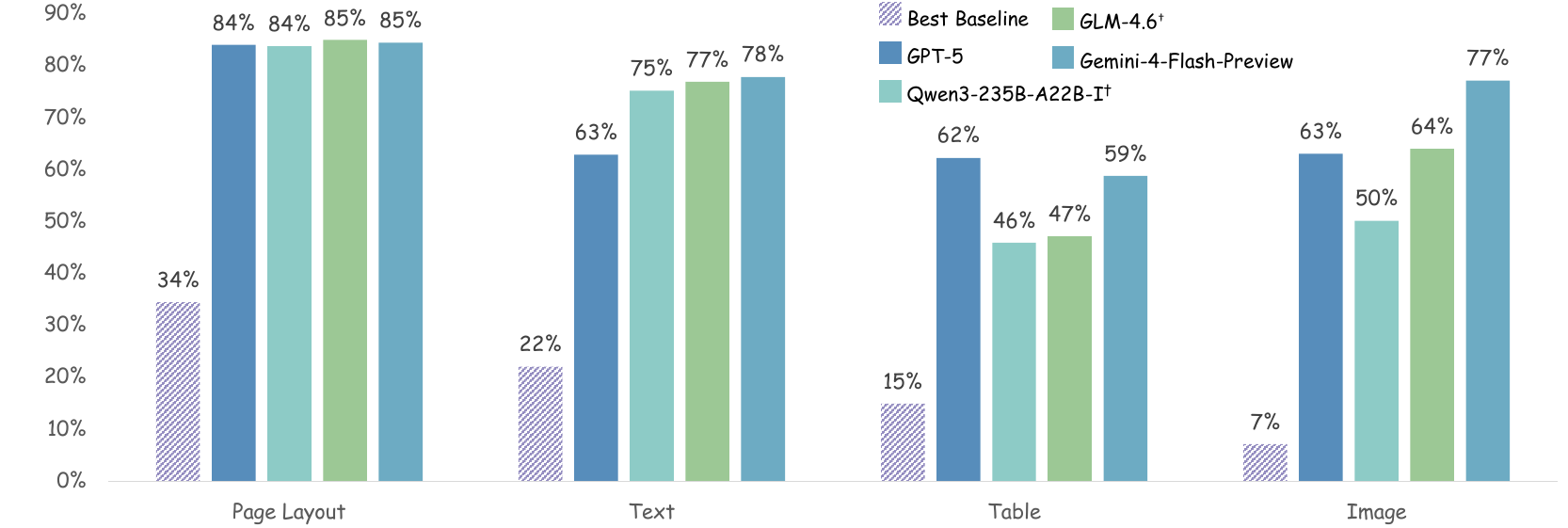}
    \caption{Formatting accuracy across target categories.}
    \label{fig:category_acc}
\end{figure}

\begin{figure}[t]
    \centering
    \includegraphics[width=0.5\textwidth]{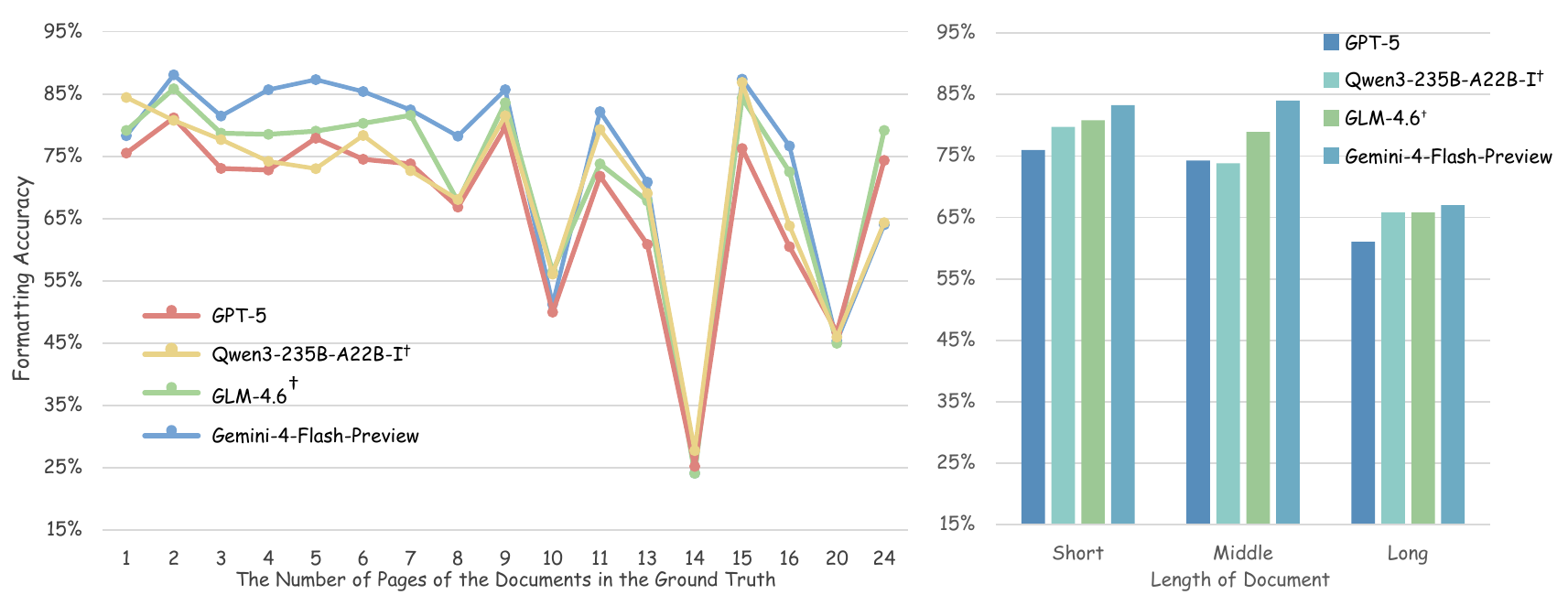}
    \caption{Impact of performance by document length.}
    \label{fig:length_Influence}
\end{figure}

We further analyze how task difficulty varies with three properties of the formatting input: formatting complexity, target category, and document length.

\paragraph{Impact of Formatting Complexity}
As shown in Figure \ref{fig:input_Influence}, we observe that formatting complexity exhibits a non-monotonic effect.
Intuitively, we hypothesize that formatting complexity increases with both the number of formatting attribute categories and the minimum number of required tool calls.
However, performance first improves and then degrades as complexity increases, peaking at moderate levels.
This suggests that both overly simple and overly complex requirements are challenging for the model.
We attribute this to sparse localization cues in simple cases.
While in complex ones, the challenge lies in precisely localizing targets among numerous candidates.

\paragraph{Impact of Target Category}
Target category results in Figure~\ref{fig:category_acc} further highlight the role of target localization difficulty.
Page adjustments perform best, as the target is usually unique and easy to locate.
Text modifications remain relatively reliable due to explicit semantic cues.
In contrast, tables and images are more challenging because their descriptions are often ambiguous and may correspond to multiple similar candidates, making precise localization difficult.
This result suggests that performance is mainly determined by how easy it is to locate the intended target.

\paragraph{Impact of Document Length}
Figure~\ref{fig:length_Influence} shows that document length has only a limited influence.
Performance remains stable across most page ranges and drops notably only on documents with 10 pages.
Manual inspection suggests that these failures mainly involve uncommon formatting requirements that the current tools cannot support.

Overall, we argue that real-world document formatting is primarily gated by accurate target localization, though tool capability may also matter for certain uncommon formatting needs.
Additional analyses on model size and verification are provided in Appendix~\ref{app:analysis}.

\section{Conclusion}

This paper focuses on the Text-to-Format task, which uses natural language instructions to automate document formatting.
We address two key limitations of current works: the absence of realistic content-aware evaluation datasets and the repeated document reviewing of ReAct-style methods in this task.
To bridge these gaps, we introduce \docformbench, a benchmark that extends Text-to-Format evaluation to diverse content-aware formatting requests. It comprises 500 real-world instances with automatic metrics for both accuracy and efficiency.
We further propose \docformflow, a four-stage method that separates \textit{what to modify} from \textit{how to modify}.
We construct extensive experiments across multiple LLMs and multimodal models, demonstrating that \docformflow consistently outperforms all baseline approaches.
Our analysis further reveals that formatting performance is primarily governed by target localization capability.
We hope that \docformbench and \docformflow provide a solid foundation for future research.

\section*{Limitations}
We summarize the limitations of this work from three aspects.

Firstly, due to computational constraints, we did not evaluate the latest large-scale proprietary models (e.g., GPT-5.5, GLM-5-Turbo), nor certain baselines with high token consumption under equivalent budgets.
We instead focus on representative open-source models, including Qwen and GLM series, which provide a consistent and informative comparison across model families.

\begin{figure}[t]
    \centering
    \includegraphics[width=0.5\textwidth]{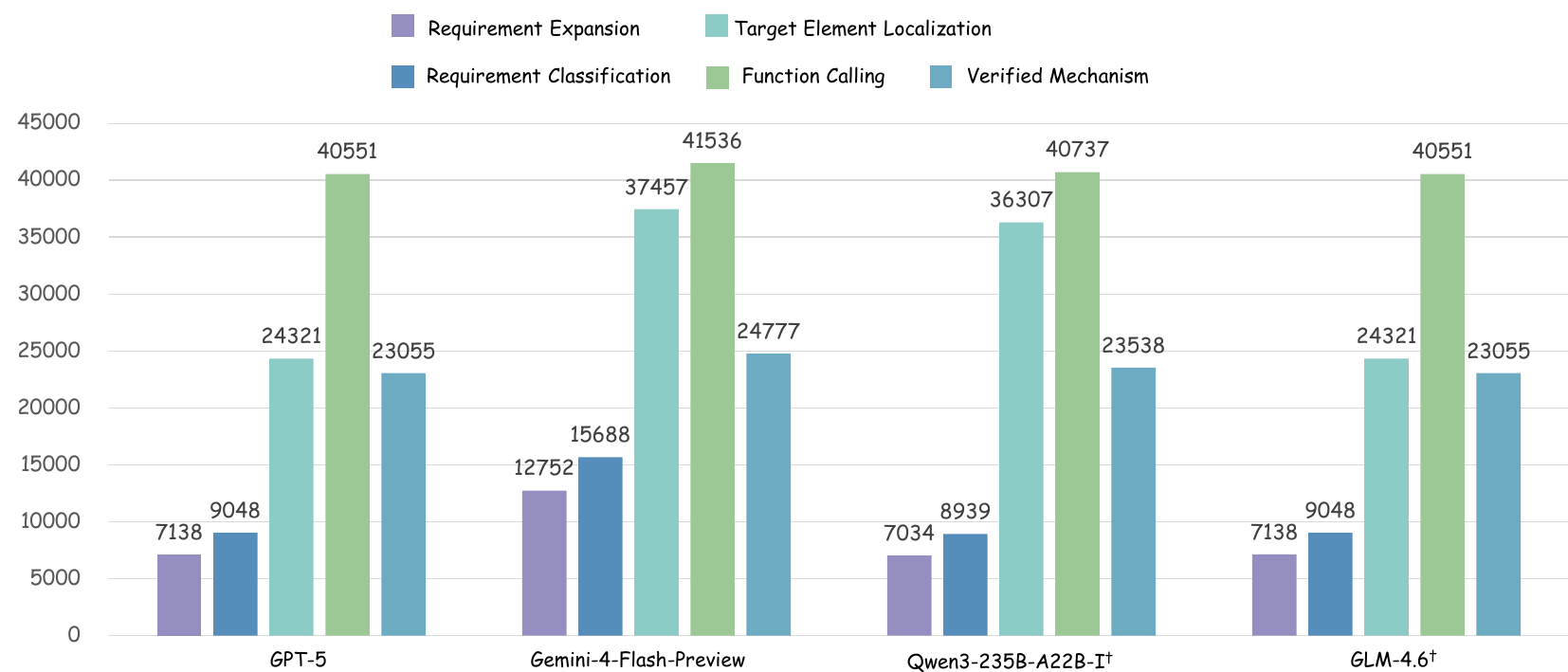}
    \caption{Token consumption of different stages for \docformflow on various LLMs.}
    \label{fig:token_consumption}
\end{figure}

Secondly, our method relies on multi-step localization and tool invocation, which introduces non-trivial inference overhead.
We analyze token consumption across different stages of \textsc{DocFormFlow}, as shown in Figure~\ref{fig:token_consumption}.
We observe that the Function Calling stage (i.e., tool retrieval) dominates overall token usage, followed by localization and verification.
This overhead is largely driven by long tool descriptions and associated prompt-level documentation required for formatting operations.
In addition, since current LLMs lack strong prior knowledge of document formatting, external tool specifications must be explicitly injected, further increasing context length.
Across four best-performing configurations, a single formatting task requires approximately 100K tokens on average.
This suggests that while multi-step reasoning improves reliability, it introduces significant inference cost.
Reducing this overhead will likely require more compact tool representations and better parameterized formatting primitives, which we leave for future work.

Finally, \textsc{DocFormFlow} is implemented on top of the pywin32 COM interface for Microsoft Word.
While this enables access to a broad set of formatting operations, the current tool abstraction does not fully cover all Word functionalities.
Advanced features such as complex shapes, floating layouts, numbering schemes, and references are only partially supported.
In addition, the system requires a Windows environment with Microsoft Word installed and is not natively cross-platform.

\section*{Ethics Statement}
This paper focuses on the Text-to-Format task and proposes an automatic formatting method, \textsc{DocFormFlow}, and an evaluation dataset, \textsc{DocFormBench}. The proposed method operates on document formatting only, without modifying the semantic content of the documents. In addition, \textsc{DocFormFlow} is built upon the COM interfaces provided by the open-source pywin32 library to interact with Microsoft Word. Therefore, the methodology does not involve any direct human subjects or sensitive decision-making scenarios.

\textsc{DocFormBench} is constructed through manual template collection, modification, and verification. Specifically, two annotators collected templates from publicly available sources, and three annotators performed review and rewriting. All annotators are graduate students and participated on a voluntary basis. All documents are sourced from publicly available data, and we carefully anonymize any potential personal information (e.g., names) to ensure privacy protection. We also filter out any content that may introduce social bias or harmful instructions.

To improve linguistic diversity and better reflect real-world formatting scenarios, the dataset includes both Chinese and English documents. All data are manually anonymized and balanced across languages to reduce potential language-specific bias during evaluation.

\bibliography{custom}
\appendix

\section{Appendix}

\subsection{Detailed Resource Examples}
\label{app:resource_details}
Each template is presented as a pair: a textual requirement and a set of configurable parameters. and the corresponding parameter fields.
The following example illustrates a Page Layout Template and a Paragraph Template for an academic paper in English.
The Page Layout Template is shown below:
\begin{center}
\begin{tcolorbox}[colframe=white,colback=gray!5, boxsep=-1pt, width=0.95\linewidth, enhanced]
\begin{minipage}{\textwidth}
\fontsize{8.5pt}{10.5pt}\selectfont\ttfamily
\textbf{\textit{Textual Requirement}}: \\
The document must be formatted for US Letter paper size (8.5''×11'') with 1.5-inch margins on all sides, single column, and no document grid. \\
\textbf{\textit{Configurable Parameters}}: \\
\{
    "margin": \\
    \{"top": 1.5 inches, \\
    "bottom": 1.5 inches, \\
    "left": 1.5 inches,\\
    "right": 1.5 inches\},\\
    "paper":\\
    \{"size": "US Letter",\\ "width": 8.5,\\ "height": 11\},\\
    "grid": \{"enabled": false\},\\
    "columns": \{"count": 1\} \}
\end{minipage}
\end{tcolorbox}
\end{center}
The Paragraph Template is shown below:
\begin{center}
\begin{tcolorbox}[colframe=white, colback=gray!5, boxsep=-1pt, width=0.95\linewidth, enhanced]
\begin{minipage}{\linewidth}
\fontsize{8.5pt}{10.5pt}\selectfont\ttfamily
\textbf{\textit{Textual Requirement}}: \\
-- Paper title: 15 pt Times New Roman bold, centered, 10 pt space after, 1.5 line spacing. \\
-- Abstract text: 10 pt Times New Roman, 12 pt space after. \\
...

\textbf{\textit{Configurable Parameters}}:
\begin{verbatim}
[
{"style_name": "PaperTitle", 
  "settings": 
    {"font": 
      {"name": "Times New Roman", 
      "size": 15, 
      "bold": true},
    {"paragraph":
      "alignment": "center", 
      "spacing":
        {"linespacing": 1.5, 
        "spaceafter": 10},}},
{"style_name": "AbstractText", 
  "settings":
    {"font":
      {"name": "Times New Roman", 
      "size": 10},
    "paragraph":
      { "spacing": 
        {"spaceafter": 12},}}
...
]
\end{verbatim}
\end{minipage}
\end{tcolorbox}
\end{center}
The Document Content is shown below:
\begin{center}
\begin{tcolorbox}[colframe=white,colback=gray!5, boxsep=-1pt, width=0.95\linewidth, enhanced]
\begin{minipage}{\textwidth}
\fontsize{8.5pt}{10.5pt}\selectfont\ttfamily
\textbf{Document Content} \\
-- PaperTitle: "Towards Robust Document Formatting Understanding" \\
-- AbstractText: "This paper introduces a benchmark for evaluating document formatting capabilities..." \\
-- ...
\end{minipage}
\end{tcolorbox}
\end{center}

\subsection{Statistics of DocFormBench}
\label{app:statistic}

\begin{figure}[t]
    \centering
    \includegraphics[width=0.5\textwidth]{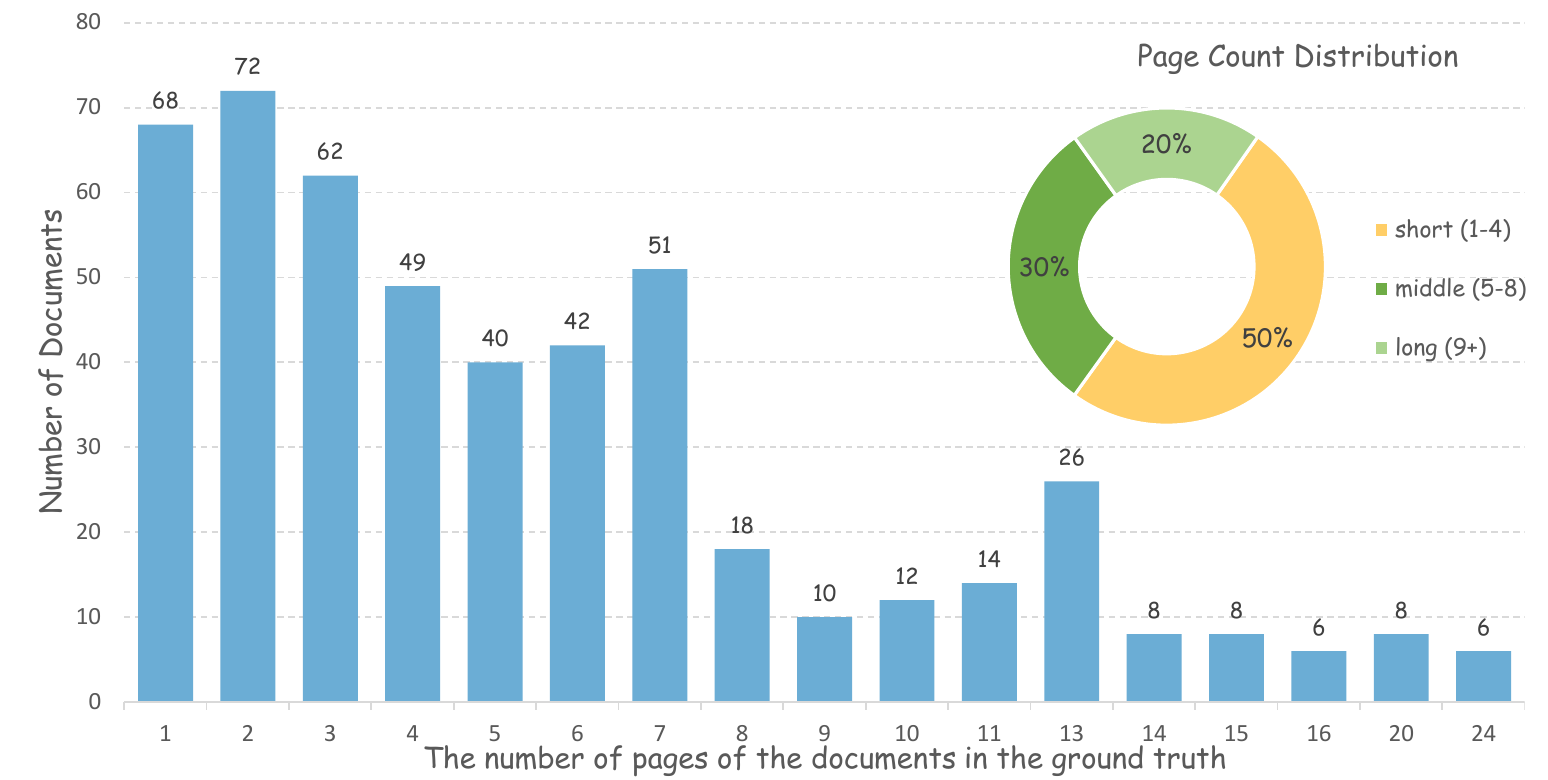}
    \caption{Length distribution of Documents in \docformbench.}
    \label{fig:length_distribution}
\end{figure}

\begin{figure}[t]
    \centering
    \includegraphics[width=0.5\textwidth]{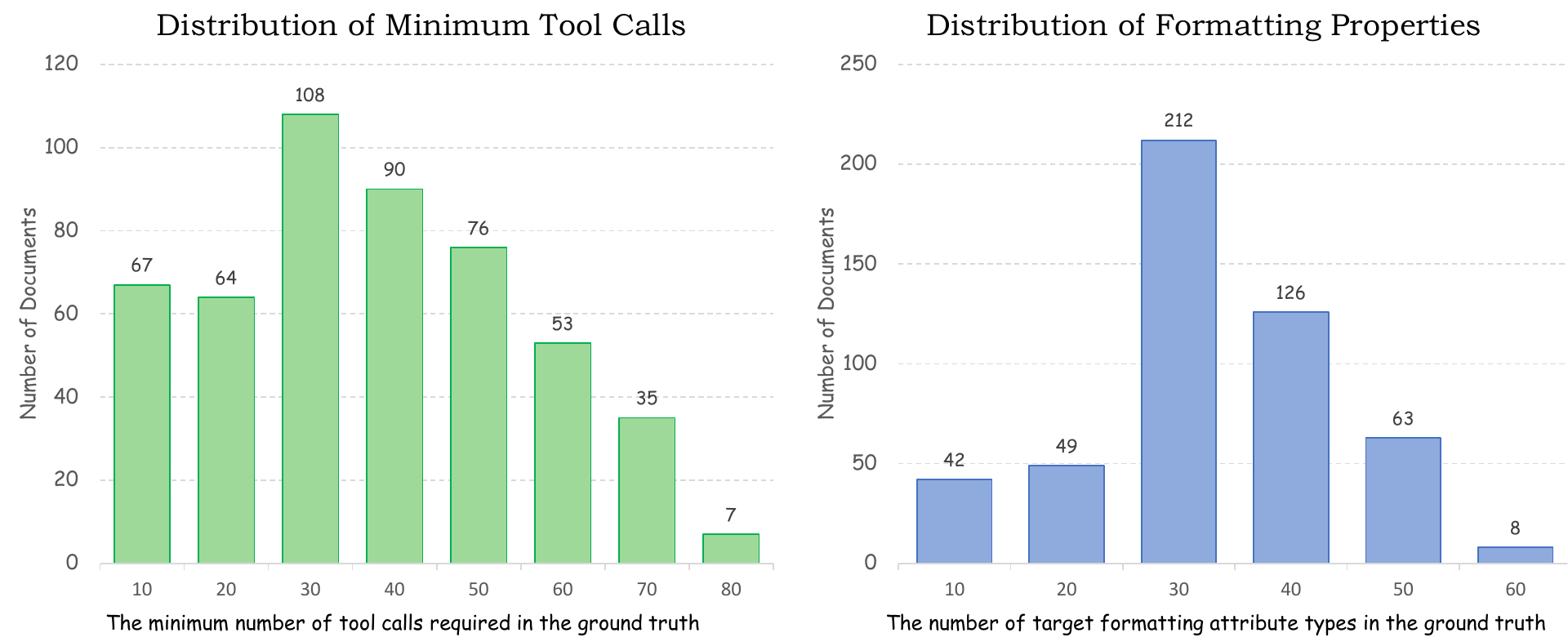}
    \caption{Complexity distribution of Formatting Requirements in \docformbench.}
    \label{fig:complex_distribution}
\end{figure}

To provide a comprehensive understanding of \docformbench, we analyze the benchmark from five aspects: the document length distribution, the input length distribution of formatting requirements, the complexity distribution of formatting adjustment tasks, the coverage of supported formatting attributes, and the distributions of document types and languages.

The document length distribution is shown in Figure \ref{fig:length_distribution}. 
We measure document length by the page count of the initial document before formatting adjustment $D_{\text{src}}$.  
The overall distribution exhibits an approximate long-tail pattern, which is consistent with real-world document scenarios.  
Following common conventions in writing practice, we categorize documents with 1--4 pages as short,  5--8 pages as medium-length,  and 9 pages or more as long.  
As illustrated in Figure \ref{fig:length_distribution}, the proportion of short, medium-length, and long documents is approximately 5:3:2.  
This distribution suggests that \docformbench\ covers documents of varying lengths while remaining aligned with realistic usage conditions,  where shorter documents are more frequent but longer documents are still sufficiently represented.  

\begin{table}[!t]
  \centering
  \resizebox{\columnwidth}{!}{%
    \begin{tabular}{ll}
      \toprule
      \textbf{Tool Name} & \textbf{Tool Description} \\
      \midrule
      
      \rowcolor{gray!20} \multicolumn{2}{l}{\textit{Page Layout}} \\
      Page Margins & Control the top, bottom, left, and right margins \\
      Header Distance & Set the distance of headers from page edges \\
      Footer Distance & Set the distance of footers from page edges \\
      Header Content & Customize header text, font, alignment, and bottom border \\
      Footer Content & Define footer text, font, alignment, and page numbering style \\
      Gutter & Adjust binding gutter width and position  \\
      Paper & Choose paper size, orientation, or define custom dimensions \\
      Columns & Configure number of columns, spacing, width, and separator lines \\
      Grid & Specify layout mode, lines per page, and characters per line \\
      \midrule
      
      \rowcolor{gray!20} \multicolumn{2}{l}{\textit{Paragraph}} \\
      Base Font & Modify font family, size, bold, italic, underline, color, and highlight \\
      Advanced Font & Apply strikethrough, subscript, superscript, caps, spacing, scaling, and more \\
      Outline Level & Assign heading levels or mark as body text \\
      Alignment & Align text left, center, right, justified, or distributed \\
      Pagination& Manage widow control, keep with next, keep together, and page breaks \\
      Spacing & Adjust line spacing and paragraph spacing \\
      Indent & Set left/right indent, first-line indent, or hanging indent \\
      \midrule
      
      \rowcolor{gray!20} \multicolumn{2}{l}{\textit{Table}} \\
      Width & Set table width as a percentage of page width \\
      Alignment & Align table horizontally and cell content vertically \\
      Pagination & Control row breaks, repeat header rows, keep with next, and page breaks \\
      Left Indent & Indent the entire table from the left margin \\
      Text Wrapping & Enable or disable text wrapping around the table \\
      \midrule
      
      \rowcolor{gray!20} \multicolumn{2}{l}{\textit{Image}} \\
      Size & Specify absolute width and height \\
      Scaling & Scale the image by a percentage relative to its current size \\
       Alignment & Align the image horizontally within its paragraph \\
      Pagination & Control keep with next, keep together, and page break before \\
      \bottomrule
    \end{tabular}
  }
  \caption{Supported Formatting Tools and Description.}
    \label{tab:formatting_attributes}
\end{table}

We analyze the complexity of formatting tasks along two intuitive dimensions: execution complexity and requirement diversity.
Execution complexity is measured by the minimum number of tool calls needed to complete an instruction, derived from the recorded formatting parameters during benchmark construction.
Requirement diversity is measured by the number of distinct formatting attribute categories involved in the target specification; more categories imply broader formatting knowledge and greater task difficulty.
As shown in Figure~\ref{fig:complex_distribution}, both metrics approximately follow a normal distribution.
This indicates that most samples require a moderate level of interaction and cover a reasonable range of formatting knowledge, while only a few cases are extremely simple or highly complex.
Such a distribution supports a benchmark with balanced difficulty and practical utility.

As shown in Table~\ref{tab:object_distribution}, the distributions of formatting object types and instruction styles are highly balanced between Chinese and English,  
indicating that \docformbench\ does not introduce an obvious language bias at this level.  
Across both languages, page and paragraph formatting constitute the majority of formatting requirements, which is consistent with realistic text-to-format scenarios, where most user requests focus on document layout and text organization.

\begin{table}[!t]
  \centering
  \resizebox{\columnwidth}{!}{%
    \begin{tabular}{llcc}
      \toprule
      \textbf{Domain} & \textbf{Document Type} & \textbf{Chinese} & \textbf{English} \\
      \midrule

      \multirow{3}{*}{Academic}
      & Research Paper    & 36 & 46 \\
      & Academic Thesis   & 18 & 14 \\
      & Scientific Poster  & 12 & 12 \\
      \midrule

      \multirow{5}{*}{Official} 
      & Legal Contract       & 24 & 27 \\
      & Examination Paper    & 24 & 24 \\
      & Project Proposal     & 24 & 20 \\
      & Government Document  & 18 & 12 \\
      & Court Judgment       & 16 & 16 \\
      \midrule

      \multirow{2}{*}{Publishing}
      & Newspaper            & 18 & 18 \\
      & Postcard             & 12 & 12 \\
      \midrule

      \multirow{2}{*}{Personal} 
      & General Note         & 36 & 37 \\
      & Personal Letter      & 12 & 12 \\
      \bottomrule

    \end{tabular}
  }
  \caption{Distribution of Dataset across Different Document Types and Languages.}
    \label{tab:type_distribution}
\end{table}

More specifically, page-level formatting is dominated by content-agnostic instructions,  
since operations such as page margins, page size, header and footer positions, and column settings typically do not depend on the semantic content of the document.  
In contrast, paragraph-level formatting is primarily content-aware, because adjustments such as alignment, indentation, spacing, or selective styling are often associated with specific textual content or structural roles within the document.  
By comparison, image and table requirements account for a relatively small proportion of the benchmark.  
This also matches practical usage, as such elements appear less frequently than plain text and page layout in general document formatting tasks.

The supported formatting attributes and tools are summarized in Table \ref{tab:formatting_attributes}.  
\docformbench\ covers a broad range of formatting operations,  
including page layout, paragraph style, table, and image.

The distributions of document types and languages are presented in Table \ref{tab:type_distribution}.  
The benchmark includes multiple document types from academic, official, publishing, and personal domains,  together with both Chinese and English samples.  
The sample counts are relatively balanced across categories, but they are not forced to be strictly uniform.  
Instead, the allocation is determined by jointly considering category diversity and the usage frequency of document types in real-world scenarios.  
As a result, \docformbench\ maintains reasonable coverage of different domains and languages  
while preserving a distribution that is closer to realistic document formatting tasks.

\begin{table}[htbp]
  \centering
  \resizebox{\columnwidth}{!}{%
    \begin{tabular}{llcc}
      \toprule
      \textbf{Object Type} & \textbf{Instruction Style} & \textbf{Chinese} & \textbf{English} \\
      \midrule

      \multirow{2}{*}{Page}
      & Content-Agnostic  & 208 & 209 \\
      & Content-Aware & 20  & 20  \\
      \midrule

      \multirow{2}{*}{Paragraph}
      & Content-Agnostic  & 24  & 24  \\
      & Content-Aware & 214 & 215 \\
      \midrule

      \multirow{2}{*}{Image}
      & Content-Agnostic  & 12  & 12  \\
      & Content-Aware & 19  & 20  \\
      \midrule

      \multirow{2}{*}{Table}
      & Content-Agnostic  & 7   & 7   \\
      & Content-Aware & 10  & 14  \\
      \bottomrule

    \end{tabular}
  }
    \caption{Distribution of Formatting Object Types and Instruction Styles.}
      \label{tab:object_distribution}
\end{table}

\subsection{Metrics Calculation Details}
\label{app:metrics}

The following metrics together provide a comprehensive assessment of each method's ability to perform accurate, precise, and efficient document formatting on \docformbench.

\subsubsection{Resource Efficiency Metric}
The following metrics measure resource efficiency:
\begin{itemize}[nosep]
    \item \textbf{LLM Calls}: the average number of LLM API calls required to complete one task.
    \item \textbf{AVG Tokens}: the average number of tokens (input + output) consumed per API call.
\end{itemize}

\subsubsection{Formatting Accuracy (FA)}

In document formatting tasks, the attributes that must be modified ($\mathcal{A}_{\text{mod}}$) are far fewer than those that must stay unchanged. We therefore evaluate formatting accuracy as \textbf{recall} on the required changes.
For each document, let $V(a)$ be the attribute value in the model's output and $V_{tgt}(a)$ the target value. Define the set of \emph{correctly modified} attributes as
\[
\mathcal{A}_{\text{correct}} = \{ a \in \mathcal{A}_{\text{mod}} \mid V(a) = V_{tgt}(a) \}.
\]
\textbf{Formatting Accuracy (FA)} is then
\begin{equation}
\text{FA} = \frac{|\mathcal{A}_{\text{correct}}|}{|\mathcal{A}_{\text{mod}}|}.
\end{equation}
High FA indicates the model successfully applies the required formatting without omissions.

Because object types (Page, Text, Image, Table) appear in highly unequal numbers, a sample's FA is obtained by first averaging FA within each object type, then averaging over all types present in the sample. This avoids having any single type dominate the score.

\subsubsection{Formatting Scope Precision}

To measure whether the model introduces side effects, we check how \emph{precise} its modifications are. Let $\mathcal{M}$ be the set of all attributes the model actually changed. The \textbf{Hallucination Formatting Rate (HFR)} denotes as HFR = $1 - \text{Precision}$, where Precision is the proportion of modifications that are both intended and correct:
\begin{equation}
\text{Precision} = \frac{|\mathcal{M} \cap \mathcal{A}_{\text{correct}}|}{|\mathcal{M}|}.
\end{equation}
If the model made no modifications ($\mathcal{M}=\emptyset$), we set Precision = 1 and HFR = 0. Lower HFR means the model rarely touches attributes outside the intended scope, i.e., fewer hallucinated alterations.

Following the same aggregation strategy as FA, per-sample HFR is computed by averaging within each object type first and then across the types present in the sample, to fairly account for imbalanced object distributions.

\paragraph{Illustrative Example}
Below is a concrete example showing how FA (recall) and HFR ($1-\text{precision}$) are computed for a single document.

\begin{center}
\begin{tcolorbox}[colframe=white,colback=gray!5, boxsep=-1pt, width=0.95\linewidth, enhanced]
\begin{minipage}{\textwidth}
\fontsize{9pt}{11pt}\selectfont\ttfamily
\textbf{Required changes} ($\mathcal{A}_{\text{mod}}$): \\
\ \ \texttt{Text1.font-size} $\to$ \texttt{12}, \ \ \texttt{Text1.bold} $\to$ \texttt{true}, \ \ \texttt{Table1.alignment} $\to$ \texttt{"center"} \\[2pt]
\textbf{Attributes outside $\mathcal{A}_{\text{mod}}$ that must stay unchanged}: \\
\ \ \texttt{Page1.margin} must remain \texttt{1 inch} \\[4pt]
\textbf{Model output}: \\
\ \ \texttt{Text1.font-size} $\to$ \texttt{12} \textcolor{green}{\checkmark}, \ 
\texttt{Text1.bold} $\to$ \textcolor{red}{\texttt{false}}, \ 
\texttt{Table1.alignment} $\to$ \texttt{"center"} \textcolor{green}{\checkmark}, \\
\ \ \texttt{Page1.margin} $\to$ \textcolor{red}{\texttt{2 inches}} \\[4pt]
\textbf{FA (Recall)}: \\
\ \ $\mathcal{A}_{\text{correct}} = \{\texttt{Text1.font-size},\; \texttt{Table1.alignment}\}$ \\
\ \ \texttt{Text}: $|\mathcal{A}_{\text{correct}}|=1$, $|\mathcal{A}_{\text{mod}}|=2$ $\to$ FA = 0.5 \\
\ \ \texttt{Table}: $|\mathcal{A}_{\text{correct}}|=1$, $|\mathcal{A}_{\text{mod}}|=1$ $\to$ FA = 1.0 \\
\ \ \texttt{Page}: no attributes in $\mathcal{A}_{\text{mod}}$ $\to$ FA = 1.0 (by definition) \\
\ \ $\text{FA}_{\text{sample}} = (0.5 + 1.0 + 1.0)/3 = \mathbf{0.833}$ \\[4pt]
\textbf{HFR ($1-\text{Precision}$)}: \\
\ \ $\mathcal{M}_{\text{Page}} = \{\texttt{Page1.margin}\}$, 
   $\mathcal{M}_{\text{Page}} \cap \mathcal{A}_{\text{correct}} = \emptyset$
   $\to$ Precision = 0, HFR = 1.0 \\
\ \ $\mathcal{M}_{\text{Text}} = \{\texttt{Text1.font-size},\; \texttt{Text1.bold}\}$, 
   $\mathcal{M}_{\text{Text}} \cap \mathcal{A}_{\text{correct}} = \{\texttt{Text1.font-size}\}$
   $\to$ Precision = $1/2$, HFR = 0.5 \\
\ \ $\mathcal{M}_{\text{Table}} = \emptyset$ $\to$ Precision = 1, HFR = 0 \\
\ \ $\text{HFR}_{\text{sample}} = (1.0 + 0.5 + 0.0)/3 = \mathbf{0.5}$
\end{minipage}
\end{tcolorbox}
\end{center}
\subsection{Models used in Experiment}
The models used in this experiment are shown in the Table \ref{tab:model_list}.
\begin{table}[!t]
  \centering
  \resizebox{\columnwidth}{!}{%
    \begin{tabular}{llcc}
      \toprule
      \textbf{Model Name} & \textbf{Mod.} & \textbf{Params} & \textbf{Thinking} \\
      \midrule
      \rowcolor{gray!20} \multicolumn{4}{l}{\textit{Closed-source Models}} \\
      Gemini-3-Flash-Preview~\citep{gemini3} & MM   & --   & S \\
      GPT-5~\citep{gpt5}                  & MM   & --   & S \\
      \midrule
      \rowcolor{gray!20} \multicolumn{4}{l}{\textit{Open-source Models}} \\
      GLM-4.6~\citep{glm46}                & Text & 355B & S \\
      GLM-4.6V               & MM   & 106B & S \\
      DeepSeek-V3.2~\citep{deepseek_v32}          & Text & 685B & S \\
      DeepSeek-V4-Flash~\citep{deepseek_v4}      & Text & 284B & S \\
      DeepSeek-V4-Pro        & Text & 1600B& S \\
      Qwen3-235B-A22B-2507-I~\citep{qwen3}\tablefootnote{I = Instruction variant, T = Thinking variant.} & Text & 235B & N \\
      Qwen3-235B-A22B-2507-T             & Text & 235B & Y \\
      Qwen3-VL-235B-A22B-2507-I           & MM   & 235B & N \\
      Qwen3-VL-235B-A22B-2507-T           & MM   & 235B & Y \\
      Qwen3-VL-32B-I              & Text & 32B & N \\
      Qwen3-VL-30B-A3B-I              & Text & 32B & N \\
      Qwen3-VL-8B-I              & Text & 8B & N \\
      Qwen3.5-35B-A3B~\citep{qwen35}               & MM   & 35B  & S \\
      Qv3.5-122B-A10B              & MM   & 122B & S \\
      Qwe3.5-397B-A17B              & MM   & 397B & S \\
      \bottomrule
    \end{tabular}
  }  
  \caption{List of supported models grouped by series. MM denotes multimodal models; Y/N/S indicate whether reasoning is enabled, disabled, or switchable.}
  \label{tab:model_list}
\end{table}
\subsection{Detailed Hyperparameter Settings}
\label{app:param_settings}

\paragraph{Loop Prevention Mechanism}
During experiments, we observe that some baseline methods may not be compatible with certain models, leading to repeated sequence operations and unnecessary computational overhead. To detect and exit such loops, we designed the following mechanism. If the tool calls and observations are identical for 5 consecutive steps, the task is forcibly terminated.
This strategy effectively prevents infinite loops while ensuring the stability of experimental results and efficient use of resources.

\paragraph{Reasoning Mode Settings}

To study the effect of reasoning, we configure each model as follows:

\begin{itemize}[nosep]
    \item \textbf{Gemini-3-Flash-Preview}: reasoning enabled by setting reasoning level to \texttt{high}.
    \item \textbf{GLM series}: reasoning enabled by setting reasoning level to \texttt{auto}.
    \item \textbf{Qwen-3 series}: use the corresponding \texttt{Thinking} variant.
    \item \textbf{GPT-5}: reasoning cannot be fully disabled. For default (minimal reasoning) mode, we set reasoning to \texttt{minimal} to approximate a disabled state; reasoning enabled mode is set to \texttt{high}.
\end{itemize}

All other models run with reasoning mode disabled by default unless explicitly studying its effect.
\subsection{Additional Experimental Analysis}
\label{app:analysis}

\subsubsection{Effect of Model Scale and Series}
\begin{table}[!t]
  \centering
  \resizebox{\columnwidth}{!}{%
    \begin{tabular}{lrrrr} 
      \toprule
      \textbf{Model Series}  & \textbf{AVG Calls} & \textbf{FA(\%)} & \textbf{HFR(\%) $\downarrow$ } & \textbf{AVG Tokens} \\
      \midrule
      \rowcolor{gray!20} \multicolumn{5}{l}{\textit{DeepSeek Series}} \\
      Deepseek-V3.2       & 34.77 & 72.87 & 1.99 & 2,802 \\
      Deepseek-V4-Flash     & 47.24 & 64.36 & 1.80 & 2,771 \\
      Deepseek-V4-Pro      & 33.08 & 74.00 & 1.57 & 3,172 \\
      \midrule
      \rowcolor{gray!20} \multicolumn{5}{l}{\textit{Qwen3 Series}} \\
      Qwen3-VL-235B-A22B-I  & 29.28 & 70.94 & 2.03 & 3,246 \\
      Qwen3-VL-30B-A3B-I  & 32.51 & 64.47 & 3.80 & 3,419 \\
      Qwen3-VL-32B-I    & 33.01 & 73.04 & 1.94 & 3,316 \\
      Qwen3-VL-8B-I     & 39.08 & 47.64 & 2.45 & 3,206 \\
      \midrule
      \rowcolor{gray!20} \multicolumn{5}{l}{\textit{Qwen3.5 Series}} \\
      Qwen3.5-35B-A3B    & 30.31 & 62.10 & 3.45 & 4,042 \\
      Qwen3.5-122B-A10B  & 33.66 & 70.87 & 3.06 & 3,259 \\
      Qwen3.5-397B-A17B  & 32.23 & \textbf{75.19} & 2.69 & 3,281 \\
      \bottomrule
    \end{tabular}
  }
  \caption{Performance Comparison of Different Model Series and Parameters.}
    \label{tab:model_comparison}
\end{table}

We conduct a scaling experiment to explore how model scale and architecture affect \docformflow's performance across three model series. The results are shown in Table~\ref{tab:model_comparison}.

\paragraph{Larger models consistently improve accuracy within each series.}
Scaling up active parameters yields clear gains: Deepseek-V4-Pro outperforms Deepseek-V4-Flash by +9.64\% FA, and Qwen3.5-397B-A17B surpasses Qwen3.5-35B-A3B by +13.09\%.
Within the Qwen3-VL series, Qwen3-VL-32B-I (32B active) achieves 73.04\% FA, substantially ahead of Qwen3-VL-30B-A3B-I (3B active) at 64.47\% and Qwen3-VL-8B-I at 47.64\%.
We argue that stronger instruction following and more precise target localization come with increased model capacity, both critical for formatting tasks.

\paragraph{Architecture matters as much as parameter count.}
Cross-series comparison reveals that parameter count alone does not determine performance.
Deepseek-V4-Flash trails Qwen3-VL-32B-I despite comparable scale, and Qwen3-VL-235B-A22B-I (22B active) underperforms Qwen3.5-397B-A17B (17B active) by over 4\% FA.
MoE-based models in the DeepSeek and Qwen3.5 series achieve competitive or superior performance with fewer activated parameters per token, offering a better efficiency--capability trade-off.

\begin{table}[!t]
  \centering
  \label{tab:verified}
  \resizebox{\columnwidth}{!}{
    \begin{tabular}{lllll}
      \toprule
      \textbf{Model} & \textbf{Check Only Fail} & \textbf{Retry Only Fail} & \textbf{Both Fail} & \textbf{Overall} \\
      \midrule
      \rowcolor{gray!20} \multicolumn{5}{l}{\textit{No verified mechanism}} \\
      GPT-5 & 71.19 & 57.15 & 62.98 & 69.54 \\
      Gemini-3-Flash-Preview & 77.03 & 73.99 & 65.13 & 75.91 \\
      Qwen3-235b-A22B-I$^\dagger$ & 71.03 & 78.54 & 69.08 & 71.84 \\
      GLM-4.6$^\dagger$ & 74.47 & 85.35 & 64.88 & 74.24 \\
      \midrule
      \rowcolor{gray!20} \multicolumn{5}{l}{\textit{with verified mechanism}} \\
      GPT-5 & 70.31\textcolor{blue}{$_{\downarrow 0.88 }$} & 63.25\textcolor{red}{$_{\uparrow 6.10 }$} & 65.44\textcolor{red}{$_{\uparrow 2.46 }$} & 69.36\textcolor{blue}{$_{\downarrow 0.18 }$} \\
      Gemini-3-Flash-Preview & 77.03 & 76.77\textcolor{red}{$_{\uparrow 2.78 }$} & 65.13 & 76.29\textcolor{red}{$_{\uparrow 0.38 }$} \\
      Qwen3-235b-A22B-I$^\dagger$ & 71.11\textcolor{red}{$_{\uparrow 0.08}$} & 82.95\textcolor{red}{$_{\uparrow 4.41}$} & 74.03\textcolor{red}{$_{\uparrow 4.95}$} & 74.55\textcolor{red}{$_{\uparrow 2.71}$} \\
      GLM-4.6$^\dagger$ & 75.15\textcolor{red}{$_{\uparrow 0.68}$} & 86.77\textcolor{red}{$_{\uparrow 1.42 }$} & 64.81\textcolor{blue}{$_{\downarrow 0.07 }$} & 74.89\textcolor{red}{$_{\uparrow 0.65 }$} \\
      \bottomrule
    \end{tabular}
  }
  \caption{Ablation study on the verification mechanism. We compare the formatting accuracy (FA\%) of \docformflow\ with and without verification across different trigger conditions. ``No verified mechanism'' disables verification entirely, while ``with verified mechanism'' enables it. Changes are marked in red (increase) or blue (decrease).}
    \label{tab:verified}
\end{table}

\paragraph{Practical formatting is within reach for open-source models.}
Qwen3-VL-32B-I reaches 73.04\% FA, close to the best proprietary models, while Qwen3.5-397B-A17B sets a new high of 75.19\%.
This confirms that open-source models have crossed the threshold for reliable private deployment, enabling sensitive document formatting without cloud API dependence.
As each generation improves capability density, the minimum model size needed for robust formatting continues to shrink.

\subsubsection{Effect of Verification Mechanism}
\label{app:verification_analysis}

\begin{figure}[t]
    \centering
    \includegraphics[width=0.5\textwidth]{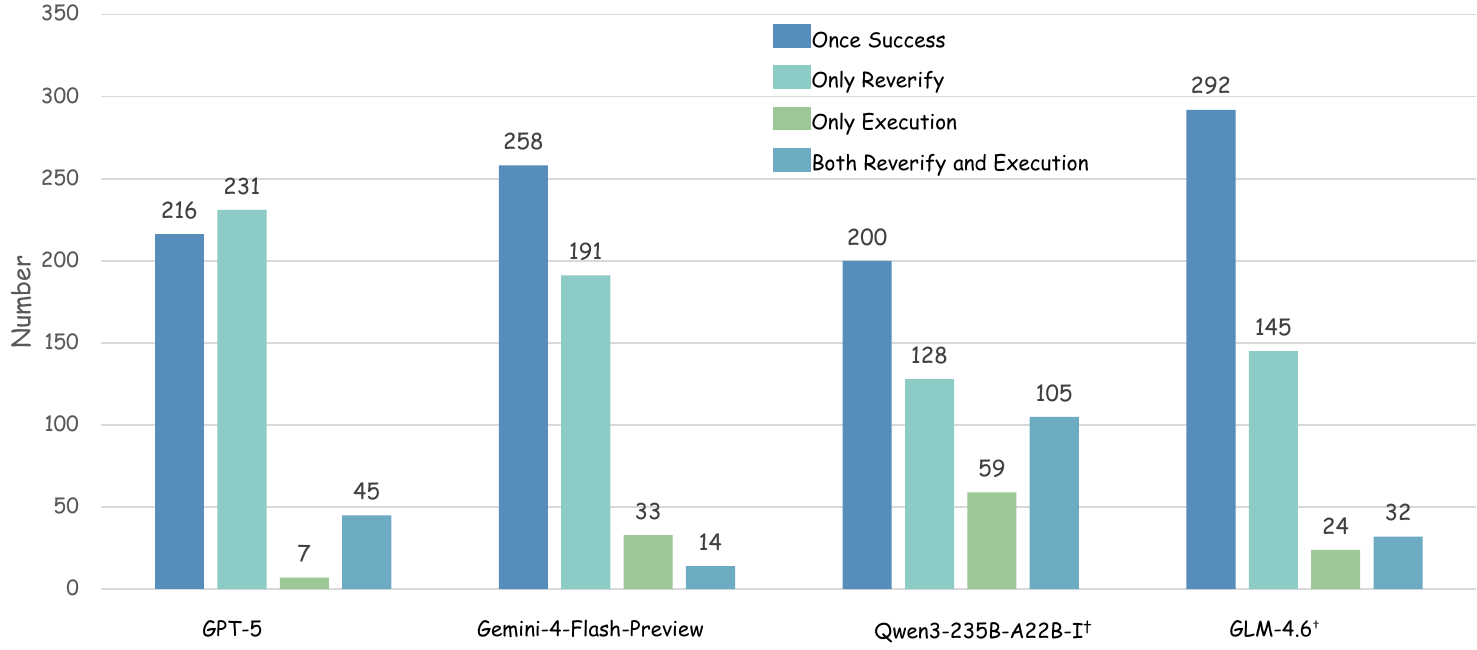}
    \caption{Distribution of Execution Retries and Verification Reverts Across Models.}
    \label{fig:retry_analysis}
\end{figure}

Figure~\ref{fig:retry_analysis} shows that most formatting tasks succeed without triggering verification. When verification is activated, it is predominantly driven by the model's own dissatisfaction with its output rather than by malformed tool calls.

Yet Table~\ref{tab:verified} reveals that verification yields only marginal gains. GPT-5 and Gemini-3-Flash-Preview see no improvement, while Qwen3-235b and GLM-4.6 gain modestly. Critically, these benefits concentrate entirely under \textit{Retry Only Fail}, which simply re-executes failed tool calls. Under \textit{Check Only Fail}, where the model re-evaluates its own output, performance is flat or slightly worse.

This sharp contrast reveals a fundamental weakness: LLMs lack the specialized formatting knowledge needed for reliable self-assessment. When asked to judge its own output, the model tends to over-adapt to the perceived user requirement, hallucinating violations in correctly formatted content and triggering unnecessary corrections. \textit{Retry Only Fail} avoids this pitfall entirely by restricting intervention to clear execution failures, making it a safer and more effective strategy.

\end{document}